\documentclass[runningheads]{llncs}
\usepackage{graphicx}
\usepackage{multirow}
\usepackage{times}
\usepackage{amsmath}
\usepackage{amssymb}
\usepackage{array}
\usepackage{color}
\usepackage{subcaption}
\usepackage{scrextend}
\usepackage{flushend}

\begin{document}
\title{On the Performance of Convolutional Neural Networks under High and Low Frequency Information}

\author{Roshan Reddy Yedla and Shiv Ram Dubey\\
\institute{Computer Vision Group,\\
Indian Institute of Information Technology, Sri City, Chittoor, Andhra Pradesh, India\\
roshanreddy.y17@iiits.in, srdubey@iiits.in}}

\maketitle

\begin{abstract}
Convolutional neural networks (CNNs) have shown very promising performance in recent years for different problems, including object recognition, face recognition, medical image analysis, etc. However, generally the trained CNN models are tested over the test set which is very similar to the trained set. The generalizability and robustness of the CNN models are very important aspects to make it to work for the unseen data. In this letter, we study the performance of CNN models over the high and low frequency information of the images. We observe that the trained CNN fails to generalize over the high and low frequency images. In order to make the CNN robust against high and low frequency images, we propose the stochastic filtering based data augmentation during training. A satisfactory performance improvement has been observed in terms of the high and low frequency generalization and robustness with the proposed stochastic filtering based data augmentation approach. The experimentations are performed using ResNet50 model over the CIFAR-10 dataset and ResNet101 model over Tiny-ImageNet dataset.\footnote{This paper is accepted in Fifth IAPR International Conference on Computer Vision and Image Processing (CVIP), 2020.}
\keywords{Convolutional Neural Networks \and Robustness \and High and Low Frequency Information \and Residual Network \and Image Classification.}
\end{abstract}

\section{Introduction}
\label{Introduction}
The emergence of deep learning has changed the way data was being handled in the early days. The existence of the large scale data sets and high end computational resources is the key to its success \cite{lecun2015deep}. Deep learning based models learn the important features from the data automatically in a hierarchical manner through the combination of linear and non-linear functions \cite{schmidhuber2015deep}. Convolutional neural network (CNN) is a special form of deep learning architecture designed to deal with the image data \cite{liu2017survey}. It consists of several layers including convolution, non-linearity, pooling, batch normalization, dropout, etc.

The AlexNet \cite{alexnet} was the revolutionary CNN model that became a state-of-the-art after winning the ImageNet Large-Scale Object Recognition Challenge in 2012 \cite{imagenet} with a great margin w.r.t. the hand-designed approaches. Motivated from the success of the AlexNet CNN model, various CNN models have been investigated for the object recognition such as VGGNet \cite{vggnet}, GoogleNet \cite{googlenet}, and ResNet \cite{resnet}, etc. The CNNs have been also used for many other applications such as image classification \cite{basha2020impact}, \cite{dubey2019diffgrad}, object localization \cite{ren2015faster}, image segmentation \cite{he2017mask}, image super-resolution \cite{tirer2019super}, image retrieval \cite{zhu2019unsupervised}, \cite{dubey2019local}, face recognition \cite{schroff2015facenet}, \cite{srivastava2019hard}, medical image analysis \cite{choi2020combining}, \cite{dubey2019local}, hyperspectral image classification \cite{roy2019hybridsn}, image to image translation \cite{babu2020pcsgan} and many more.

Though several CNN models have been investigated, the generalization and robustness of the trained model over unseen data are still a concern. Generally, the datasets are divided into two sets including a train set for the training of the CNN models and a test set to judge the performance of trained model. Various attempts have been made to generalize the CNN models by introducing the regularization layers like batch normalization \cite{batchnorm}, dropout \cite{dropout}, data augmentation \cite{perez2017effectiveness}, \cite{salamon2017deep}, ensemble generalization loss \cite{choi2020combining}, etc. The effect of different aspects of CNN is also analyzed by different researchers in the recent past, such as kernel size and number of filters of convolution layer \cite{agrawal2020using}, fully connected layers \cite{basha2020impact}, activation functions \cite{hayou2018selection}, loss functions \cite{srivastava2019performance}, etc.

In the usual dataset setup, the test set is very similar to the train and validation sets. Thus, the generalization and robustness of the trained CNN models are not tested effectively. Very limited work has been done so far where CNN meets with filtering. The median filtering based attack on images is dealt with residual dense neural network \cite{tariang2019robust}. Deep learning is also explored for median filtering forensics in the discrete cosine transform domain \cite{zhang2020deep}. The correlation filters are also utilized with CNNs for visual tracking problem \cite{ma2016correlation}. The high-frequency refinement is applied in the CNN framework for sharper video super-resolution \cite{singh2020high}. However, the effect of filtering over data is not well analyzed over the performance of CNNs.

In this letter, we study the performance of CNN models over the high and low frequency images. The basic idea originates from the thought process that if humans can identify the objects from the high and low frequency components, then the CNN models should be also able to do.
We also propose a stochastic filtering based data augmentation technique to cope with the generalization and robustness issues of CNN models for high and low frequency information. The application of this work is under scenario, where low and high frequency images are generated, including medical and hyperspectral images.

The rest of this letter is organized in following manner: Section II presents the datasets used along with the low and high frequency image datasets created; Section III presents the CNN model and training settings used in this work; Section IV illustrates the experimental results and analysis including stochastic filtering based data augmentation; and finally, Section V concludes the letter with summarizing remarks of this study.

\begin{figure*}[]
\centering
    \begin{subfigure}{0.8\columnwidth}
        \centering
        \includegraphics[width=0.98\columnwidth]{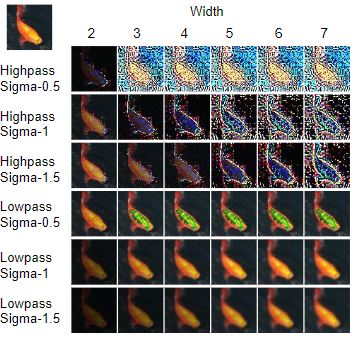}
        \caption{Example image taken from Fish category.}
        \label{fig:filtering_example1}
    \end{subfigure} \\
    \begin{subfigure}{0.8\columnwidth}
        \centering
        \includegraphics[width=0.98\columnwidth]{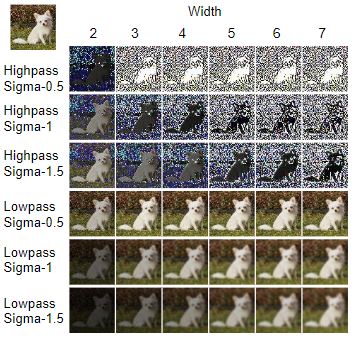}
        \caption{Example image taken from Dog category.}
        \label{fig:filtering_example2}
    \end{subfigure}
    \caption{The example high and low frequency generated images by varying the standard deviation (Sigma) and width of the kernel for filtering.}
    \label{fig:filtering_examples}
\end{figure*}

\section{High and Low Pass Image Dataset Preparation}
\label{dataset}
In order to show the effect of high and low frequency components over the performance of CNN, we use two widely adapted image classification datasets, namely, CIFAR10 \cite{krizhevsky2009learning} and TinyImageNet \cite{le2015tiny}. The CIFAR10 dataset\footnote{https://www.cs.toronto.edu/~kriz/cifar.html} consists of 50,000 images for training and 10,000 images for testing equally contributed from 10 categories.
The TinyImageNet dataset\footnote{https://www.kaggle.com/c/tiny-imagenet} contains 1,00,000 images for training and 10,000 images from testing equally distributed in 200 categories.

\subsection{New Test Sets Generation using High and Low Pass Filtering}
We generate the multiple test sets using the original test set of both datasets in high and low frequency domain to judge the robustness of CNN models.
The Gaussian filtering of the images is performed in both highpass and lowpass domains under different settings. Basically, we generate multiple high and low frequency images by varying the standard deviation ($\sigma$) and kernel width ($k$) of the filter. 

A high pass filter tends to retain the high frequency information in the image while reducing the low frequency information. The high frequency information of an image lies in its edges. Thus, the high pass filtering of an image produces the resultant image with only the edge information, or with the edges as more prominent. The lack of low frequency information makes the object recognition, challenging for the trained CNN models. 
A low pass filter tends to retain the low frequency information in the image while reducing the high frequency information. The low frequency information of an image lies in the smoother regions. Thus, the low pass filtering enforces the image to become blurry by smoothing the edge information. 
Due to the this loss of information the trained CNN model might find it hard to classify the image. 

For both high and low pass filtering over the test set, we use three different sigma values (standard deviation) for the kernel, i.e., 0.5, 1 and 1.5 and six different kernel widths including 2, 3, 4, 5, 6 and 7. Basically, we generate 36 new test sets for a dataset from the original test set that includes 18 having high frequency information and another 18 having low frequency information. The number of samples in any generated test set is same as the original test set.
The example generated images for two examples under different settings are illustrated in Fig. \ref{fig:filtering_examples}. It can be noticed that the high frequency images consists of mostly edge information, whereas the low frequency images consists of the blurry ambiance information.


\subsection{Training Set Augmentation using Stochastic Filtering}
To demonstrate the robustness problem of the CNN models against high and low pass filtered images, first we train the CNN model on the actual training sets of different datasets. However, we test the performance of the trained CNN model using the original test sets as well as the generated test sets having the transformed images.

Further, in order to improve the robustness of CNN models against the high and low frequency information of the image, we propose to augment the dataset with stochastic filtering. It is performed by transforming each image in the training set with either high pass filter or a low pass filter (selected randomly), with a random standard deviation chosen in the range of [0.25, 1.75] and random kernel size taken from the set \{2,3,4,5,6,7\}. Basically, the training set gets doubled by including the stochastic filtering based transformed images with the original images.


\begin{figure*}[!t]
    \centering
    \includegraphics[width=1\textwidth]{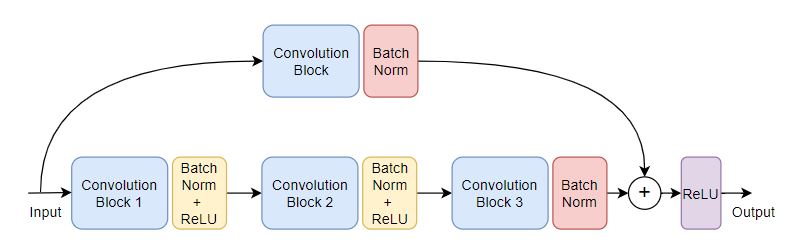}
    \caption{A bottleneck residual block of ResNet50 and ResNet101 architectures \cite{resnet}. Batch Norm refers to the batch normalization regularizer \cite{batchnorm} and ReLU refers to the rectified linear unit \cite{imagenet} based activation function.}
    \label{fig:resnet}
\end{figure*}

\section{Network Architecture and Training Settings}
This section is devoted to the description of the network architecture and training settings used for the experiments.

\subsection{Residual Network}
The ResNet architectures \cite{resnet} have been widely used for various applications including object recognition, image synthesis, image segmentation, action prediction, etc. with state-of-the-art performance \cite{khan2019survey}. In this letter, we also opt for the ResNet architectures.
We train the ResNet50 from scratch for the CIFAR-10 dataset and fine-tune the ResNet101 for the TinyImagenet dataset. The TinyImagenet dataset is having more training images and more number of classes (i.e., wider) than CIFAR10 dataset, thus, we choose the deeper ResNet for the TinyImageNet dataset. Both ResNet50 and ResNet101 use the bottleneck based residual module as illustrated in Fig. \ref{fig:resnet}. A residual block consists of the two paths, one having three convolution layers while the other one having a bottleneck convolution layer. The batch normalization \cite{batchnorm} is used for increasing the generalization of the network. The rectified linear unit (ReLU) is used as the activation function. 
The plain deeper networks are difficult to converge during training due to vanishing/exploding gradient problem.
The residual connection of ResNet architectures helps in the gradient descent optimization by flowing the gradient during backpropagation.


\subsection{Training Settings}
We have carried out all our experiments using the GPU service provided in the free tier of Google Colab. The PyTorch deep learning library is used to run the experiments. The stochastic gradient descent (SGD) optimizer is used to train the ResNet models with softmax based cross-entropy loss function with a batch size of 128. The multi-step learning rate scheduler is used with initial learning rate set at 0.1. The ResNet50 model is trained from scratch over the CIFAR10 dataset for 200 epochs with learning rate drops by a factor of 0.1 at the 100th and 150th epoch. The dimension of CIFAR10 images is $32 \times 32 \times 3$.
The pre-trained ResNet101 model is fine-tuned over the TinyImageNet dataset for 20 epochs with learning rate drops by a factor of 0.1 on every 5th epoch. The first 7 layers of ResNet101 are frozen. The TinyImageNet images are resized to $224 \times 224 \times 3$ using bicubic interpolation to make it fit with the pretrained ResNet101.
Based on the training dataset, four experiments are performed, including 1) original CIFAR10 train set, 2) original TinyImageNet train set, 3) original + stochastic filtering transformed CIFAR10 train set, and 4) original + stochastic filtering transformed TinyImageNet train set. However, the original test set as well as 36 low and high pass transformed test sets for each dataset are used to evaluate the performance in each experiment.

\begin{table}[!t]
\centering
\setlength{\tabcolsep}{0.03\columnwidth}
\caption{Testing accuracy in \% using high and low pass test sets of CIFAR-10 with actual train set. The accuracy over the original test set is 94.95\%.}
\begin{tabular}{|c|c|c|c|c|c|c|c|}
\hline
\multirow{2}{*}{Filter} & kernel & \multicolumn{6}{c|}{kernel Width} \\ \cline{3-8}
& Sigma & 2 & 3 & 4 & 5 & 6 & 7 \\ \hline
\multirow{3}{*}{High Pass} & 0.5 & 58.13 & 17.59 & 17.77 & 17.94 & 17.94 & 17.94 \\ \cline{2-8}
& 1 & 90.64 & 45.72 & 29.88 & 17.13 & 16.02 & 14.04 \\ \cline{2-8}
& 1.5 & 93.63 & 85.45 & 62.77 & 33.15 & 26.63 & 22.01 \\ \hline
\multirow{3}{*}{Low Pass} & 0.5 & 93.89 & 91.00 & 90.87 & 90.95 & 90.95 & 90.95 \\ \cline{2-8}
& 1 & 63.42 & 40.79 & 33.22 & 25.85 & 25.99 & 25.95 \\ \cline{2-8}
& 1.5 & 25.48 & 23.71 & 20.07 & 16.86 & 16.34 & 16.14 \\ \hline
\end{tabular}
\label{table:cifar_original}
\end{table}

\section{Experimental Results and Discussion}
We illustrate the classification results under different experimental settings to present the insight about the CNN robustness against high and low pass filtering.

\subsection{Experiment-1}
The original CIFAR-10 train set is used to train the ResNet50 in this experiment. The most commonly used data augmentations on the training data are used, including random cropping, horizontal flipping and normalization. The testing is conducted using original test set and the generated test sets with transformed images. The classification accuracy over \textbf{original test set} is observed as \textbf{94.95\%}. The results of the transformed test sets are summarized in Table \ref{table:cifar_original}. As expected, the performance degrades for higher values of kernel width in both the cases of high and low frequency images. The higher values of kernel width lead to the blurring in the low frequency and finer edges in high frequency. Moreover, it is also noticed that the performance of ResNet50 degrades more for smaller kernel standard deviation (i.e., Sigma) for high frequency. However, the same is also observed for low frequency, but using higher kernel standard deviation. Basically, the higher kernel width and smaller kernel sigma based high frequency fools CNN more. Similarly, the higher kernel width and higher kernel sigma based low frequency generates more confusing images for CNN.


\subsection{Experiment-2}
This experiment is performed by fine-tuning the pretrained ResNet101 on the original TinyImagenet train set and testing over original and generated test sets. Minor data augmentation such as horizontal flipping and normalization is applied to the training data. 
An accuracy of \textbf{71.41\%} is achieved on the \textbf{original validation set}. The results using the high and low pass filtering based transformed test sets are presented in Table \ref{table:tinyimagenet_original}. A similar trend of CNN robustness is followed over TinyImageNet dataset also. The performance of ResNet101 degrades heavily with higher kernel width and lower kernel sigma based high frequency images. Similarly, its performance also gets deteriorated with higher kernel width and higher kernel sigma based low frequency images.

\begin{table}[!t]
\setlength{\tabcolsep}{0.03\columnwidth}
\caption{Testing accuracy in \% using high and low pass test sets of TinyImageNet with actual train set. The accuracy over the original test set is 71.41\%.}
\begin{tabular}{|c|c|c|c|c|c|c|c|}
\hline
\multirow{2}{*}{Filter} & kernel & \multicolumn{6}{c|}{kernel Width} \\ \cline{3-8}
& Sigma & 2 & 3 & 4 & 5 & 6 & 7 \\ \hline
\multirow{3}{*}{High Pass} & 0.5 & 6.36 & 1.99 & 2.00 & 2.05 & 2.05 & 2.05 \\ \cline{2-8}
& 1 & 32.05 & 5.31 & 3.74 & 1.94 & 1.71 & 1.42 \\ \cline{2-8}
& 1.5 & 47.13 & 22.20 & 10.97 & 5.18 & 4.05 & 2.72 \\ \hline
\multirow{3}{*}{Low Pass} & 0.5 & 65.69 & 59.42 & 59.60 & 59.14 & 59.13 & 59.13 \\ \cline{2-8}
& 1 & 40.66 & 46.36 & 45.54 & 43.57 & 43.78 & 43.62 \\ \cline{2-8}
& 1.5 & 23.99 & 32.19 & 31.95 & 28.95 & 27.29 & 25.51 \\ \hline
\end{tabular}
\label{table:tinyimagenet_original}
\end{table}

\begin{table}[!t]
\setlength{\tabcolsep}{0.03\columnwidth}
\caption{Testing accuracy in \% using high and low pass test sets of CIFAR10 with stochastic filtering based augmented train set. The accuracy over the original test set is 93.94\%.}
\begin{tabular}{|c|c|c|c|c|c|c|c|}
\hline
\multirow{2}{*}{Filter} & kernel & \multicolumn{6}{c|}{kernel Width} \\ \cline{3-8}
& Sigma & 2 & 3 & 4 & 5 & 6 & 7 \\ \hline
\multirow{3}{*}{High Pass} & 0.5 & 90.82 & 75.87 & 62.22 & 77.14 & 77.14 & 63.33 \\ \cline{2-8}
& 1 & 92.98 & 89.90 & 87.98 & 86.28 & 85.76 & 84.87 \\ \cline{2-8}
& 1.5 & 93.27 & 92.43 & 90.99 & 87.48 & 87.24 & 84.41 \\ \hline
\multirow{3}{*}{Low Pass} & 0.5 & 93.11 & 92.60 & 92.60 & 92.65 & 92.65 & 92.65 \\ \cline{2-8}
& 1 & 92.18 & 90.75 & 90.34 & 90.00 & 90.01 & 89.86 \\ \cline{2-8}
& 1.5 & 91.56 & 90.25 & 89.02 & 85.35 & 84.44 & 82.91 \\ \hline
\end{tabular}
\label{table:cifar10_augmented}
\end{table}

\subsection{Experiment-3}
In order to improve the robustness of the trained CNN model for the high and low frequency information of the image, this experiment is conducted to double the CIFAR10 training set with stochastic filtering process. For each image of training set, a new image is generated by applying the high or low pass filtering (decided randomly) with random kernel width and kernel standard deviation. Note that after stochastic filtering based training data augmentation, the performance of ResNet50 over \textbf{original test set} is observed as \textbf{93.94\%} which is marginally reduced than the without augmentation. However, great performance improvement has been reported over the high and low frequency test sets as summarized in Table \ref{table:cifar10_augmented}. It shows the increased robustness of trained model w.r.t. the high and low pass images.

\begin{table}[!t]
\setlength{\tabcolsep}{0.03\columnwidth}
\caption{Testing accuracy in \% using high and low pass test sets of TinyImageNet with stochastic filtering based augmented train set. The accuracy over the original test set is 70.00\%.}
\begin{tabular}{|c|c|c|c|c|c|c|c|}
\hline
\multirow{2}{*}{Filter} & kernel & \multicolumn{6}{c|}{kernel Width} \\ \cline{3-8}
& Sigma & 2 & 3 & 4 & 5 & 6 & 7 \\ \hline
\multirow{3}{*}{High Pass} & 0.5 & 29.62 & 21.05 & 21.38 & 21.33 & 21.33 & 21.33 \\ \cline{2-8}
& 1 & 52.37 & 35.37 & 33.42 & 27.45 & 26.13 & 23.73 \\ \cline{2-8}
& 1.5 & 58.59 & 47.30 & 40.13 & 34.59 & 32.49 & 28.78 \\ \hline
\multirow{3}{*}{Low Pass} & 0.5 & 66.38 & 63.67 & 64.14 & 63.65 & 63.65 & 63.66 \\ \cline{2-8}
& 1 & 56.85 & 61.22 & 60.95 & 59.99 & 60.30 & 59.99 \\ \cline{2-8}
& 1.5 & 47.23 & 54.03 & 54.52 & 52.76 & 51.29 & 49.71 \\ \hline
\end{tabular}
\label{table:tinyimagenet_augmented}
\end{table}

\subsection{Experiment-4}
In this experiment, the stochastic filtering based data augmentation is performed over the training set of TinyImageNet dataset. Thus, doubling the number of images for training. 
The ResNet101 is fine-tuned over the transformed TinyImageNet dataset and tested over original and high and low frequency test sets. We observe an accuracy of \textbf{70.00\%} on the \textbf{original test set} which is marginally lower than the accuracy without stochastic filtering augmentation. However, a significantly improved performance is portrayed over high and low frequency test sets as described in Table \ref{table:tinyimagenet_augmented}.


\section{Conclusion}
\label{conclusion}
In this letter, we posed the robustness problem of trained CNN models for high and low frequency images. We observed that the CNN models trained over normal dataset are robust for individual low and high frequency components. It is revealed that CNN faces high difficulty to recognize the (a) high frequency images generated using higher kernel width and lower kernel standard deviation based filtering, and (b) low frequency images generated using higher kernel width and higher kernel standard deviation based filtering. In order to improve the performance of CNNs over high and low frequency components, a stochastic filtering based data augmentation approach is very useful. Very promising results are observed using the proposed stochastic filtering based data augmentation in terms of the CNN's robustness against low and high frequency components without degrading much the performance over normal images.

\bibliographystyle{splncs04}
\bibliography{References}

\end{document}